# Cyclegan Network for Sheet Metal Welding Drawing Translation

Zhiwei Song[a], Hui Yao[a] [*], Dan Tian[a], Gaohui Zhan[a]

*a Xi'an Technological University, School of Mechatronic Engineering,710000, Xi'an, Shaanxi, China.*
*songwei9023@163.com; yaohui@xatu.edu.cn; 1394121483@qq.com; 1912740549@qq.com*

**Abstract**

In intelligent manufacturing, the quality of machine translation engineering drawings will directly affect its manufacturing accuracy. Currently, most of the work is manually translated, greatly reducing production efficiency. This paper proposes an automatic translation method for welded structural engineering drawings based on Cyclic Generative Adversarial Networks (CycleGAN). The CycleGAN network model of unpaired transfer learning is used to learn the feature mapping of real welding engineering drawings to realize automatic translation of engineering drawings. U-Net and PatchGAN are the main network for the generator and discriminator, respectively. Based on removing the identity mapping function, a high-dimensional sparse network is proposed to replace the traditional dense network for the Cyclegan generator to improve noise robustness. Increase the residual block hidden layer to increase the resolution of the generated graph. The improved and fine-tuned network models are experimentally validated, computing the gap between real and generated data. It meets the welding engineering precision standard and solves the main problem of low drawing recognition efficiency in the welding manufacturing process. The results show. After training with our model, the PSNR, SSIM and MSE of welding engineering drawings reach about 44.89%, 99.58% and 2.11, respectively, which are superior to traditional networks in both training speed and accuracy.

*Keywords*：Engineering Drawings; Deep Learning; CycleGAN; Image Translation; Sparse Algorithm

## 1. INTRODUCTION

Generative Adversarial Net (GAN) networks were proposed by Goodfellow et al[1]. Inspired by zero-sum games in game theory, it treats the problem of generative recognition as a game of discriminator and generator. Generative adversarial networks consist of two deep learning networks, a generator, and a discriminator.

Nowadays, the GAN network has been applied in 12 significant fields[2], such as medicine, finance, geography, biology, etc. its network performance is relatively excellent, especially in remote sensing line recognition and style transfer; it has characteristics of high precision. Used the GAN network to identify and segment remote sensing images with high resolution and accuracy. In terms of digital line recognition, JianBo Guo, XiangRui Xu, XuDong Mao, HuiQing Qi[3][4][5][6], and others used generative adversarial networks for recognition. Their recognition effect and accuracy requirements have a high guarantee. XuChen Zhen, Hua Xu and LeFei Zhang et al[7][8][9].

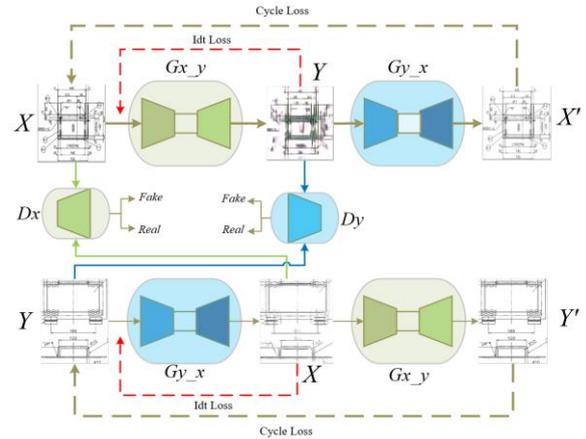

Fig 1. Two generators, *Gx_y*, *Gy_x*, and two discriminators, Dx, Dy. The real X inputs *Gx_y*, and the generator generates images in the Y domain. The resulting Y domain image is input to *Gy_x*, resulting in image X'. Compare X' and X loss. The discriminator is used to judge whether the data is true or false.

Due to the high precision and high resolution of generative adversarial networks, this paper proposes the idea of translating welding structural engineering drawings in the field of heavy industry equipment manufacturing based on the Cyclegan network. GAN network has more vital feature learning ability and generation ability than ordinary network model. Its training process is more complex and prone to gradient disappearance and model collapse[10]. GAN training instability problem, researchers, propose their solutions. Originally known as the improved DCGAN model[11], it relies on experimentally enumerating the discriminator and generator architectures and finding a better network architecture setting, which ultimately improves the instability of the GAN model training. The excellent GAN model ACGAN[12] that has appeared recently is also an improvement and optimization for the unstable training of adversarial networks. Later appeared WGAN which replaced JS and KL divergence[3][13] with EM[14], which completely solved the instability problem of GAN training. Referring to the gradient penalty method in WGAN-GP[15] can improve the training effect and speed up the convergence speed. Before this. GAN models are trained on paired datasets during style transfer and image translation training. Training on unpaired datasets. J. Zhu[16] proposed a cycle-consistent adversarial network (CycleGAN) based on GAN. It introduces cycle consistency loss for unsupervised training on the dataset, which can better achieve graph translation. This paper is based on a recurrent generative adversarial network approach to translating welded structural engineering drawings. We optimize the traditional model to achieve the desired pattern recognition effect.

---

[*] Corresponding Author

Our contributions to this paper are as follows：
- This paper applies the Cyclegan network to the welding engineering manufacturing process for the first time. Based on the cycle-consistency loss function[16] in Cyclegan, the translated welding engineering drawings can be of higher resolution, meeting industrial manufacturing accuracy standards.
- Improve the Cyclegan network. Because welding engineering drawings are grayscale graphics, we first remove the color-sensitive identity mapping loss[17].
- We add a residual block[18] to the intermediate generator layer. Improves the feature extraction capability of the generator, which can be used to generate higher-resolution graphics.
- To avoid the situation where the training speed is reduced due to the addition of intermediate layers. We apply sparse convolution and sparse activation layers K-Winners[19] instead of ReLU[20] activation layers in traditional Cyclegan generators. This speeds up training while improving the noise robustness of the generator.
- We still use Patchgan[21] as the discriminator. The Patchgan discriminator can maintain ultra-high resolution and sharpness in graphics translation. We use the discriminator's random correction unit RReLU[22] to replace the traditional Leaky Rectified Linear[23]. RReLU was first proposed and used in the Kaggle NDSB competition because $a_{ij}$ is randomly drawn from a uniform distribution during training, which can make the discriminator more stable during training.

## 2. RELATED WORK

### 2.1 GAN For Engineering Drawings

The cycle-consistent generative adversarial network adds two other loss functions to the traditional GAN network: the identity map loss and the cycle-consistency loss. The role of the identity mapping loss is to preserve the colour consistency of the generator input and output samples. In the implementation, this is done by training the K-step discriminator and training the one-step generator. Input X to the generator $Gx\_y$. It ensures that $Gx\_y$ can generate samples with a minor difference from the true distribution of the Y domain. It also ensures that when the sample input to the generator $Gx\_y$ is Y, the samples generated by the generator are as close as possible to the original Y domain distribution.

The cycle consistency loss ensures that the difference between the output sample X'(Y') restored by the cycle generation network and the original input sample X(Y) is as small as possible. The generator $Gx\_y$ learns the mapping from the X domain to the Y domain. Input X samples to $Gx\_y$; it can generate similar samples to the Y domain. $Gy\_x$ learns the mapping from the Y domain to the X domain. Input Y samples to $Gy\_x$. It can generate similar samples to the X domain. $Gx\_y$ and $Gy\_x$ are a pair of recurrent networks. The $Gy\_x$ restores the Y data generated by $Gx\_y$ to a sample X' similar to the X distribution in Fig 1. $Dx$ and $Dy$ are the discriminators of the X and Y domains, which can judge whether the generated image belongs to this domain. The U-Net network[24] is used as the main network architecture for the generator.

The intermediate generator layer adopts[18] residual network, which can increase the adaptive depth of the training network layers. The input to the generator is not noise but image samples. Input X into the generator $Gx\_y$, which can map the X-domain sample distribution to the Y-domain to achieve image-to-image translation. Using PatchGAN as a classifier[21], It can perform multi-layer convolution on the samples input to the discriminator. The PatchGAN discriminator has fewer parameters than the full-image discriminator. It can handle images of any size in a fully convolutional fashion.

### 2.2 Function

$\mathbb{E}_{x \sim p_{data}(x)}$ is the real data. $\mathbb{E}_{z \sim p_z(z)}$ is the random noise. The discriminator $D$ maximizes the probability that the input data $x$ is real or generated $G(z)$. The generator $G$ uses $[\log(1 - D(G(z)))]$ to generate the fake data with the smallest gap between the real sample data. Generating adversarial minimax objective functions:

$$\min_G \max_D V(D,G) = \mathbb{E}_{x \sim p_{data}(x)}[\log D(x)]$$
$$+ \mathbb{E}_{z \sim p_z(z)}[\log(1 - D(G(z)))] \quad (1.1)$$

Input X to the generator $Gx\_y$, which can generate fake samples Y with a distribution similar to real ones. Input the fake sample Y to another generator $Gy\_x$, and the generator, $Gy\_x$, restores the fake sample Y to sample X'. Two generators, $Gx\_y$ and $Gy\_x$, complete a loop, which enables image translation between unpaired datasets. $\mathcal{L}_{cyc}$ Indicates the $L1$ norm. Cycle consistency loss function:

$$\mathcal{L}_{cyc}(Gx\_y, Gy\_x) = \mathbb{E}_{x \sim p_{data}(x)}[\|Gy\_x(Gx\_y(x)) - x\|_1]$$
$$+ \mathbb{E}_{y \sim p_{data}(y)}[\|Gx\_y(Gy\_x(y)) - y\|_1] \quad (1.2)$$

The minimax objective function of the generative adversarial network is generated by Eq.(1.1) inference cycle consistency:

$$\mathcal{L}_{GAN}\min_{Gx\_y}\max_{D_Y}(Gx\_y,D_Y,X,Y) = \mathbb{E}_{y\sim p_{data}(y)}[\log D_Y(y)]$$
$$+ \mathbb{E}_{x\sim p_{data}(x)}[\log(1 - D_Y(Gx\_y(x)))] \quad (1.3)$$

*Idt Loss* mainly prevents the network from changing the color difference between X and Y domains in learning feature mapping. In Fig 1. the sample Y generated by *Gx_y* is forwarded to *Gx_y* again. The purpose is to judge the difference between the first generated sample Y and the second pass generated Y. The identity map loss function is as follows:

$$\mathcal{L}_{identity}(Gx\_y,Gy\_x) = \mathbb{E}_{y\sim p_{data}(y)}[\|Gx\_y(y) - y\|_1]$$
$$+ \mathbb{E}_{x\sim p_{data}(x)}[\|Gy\_x(x) - x\|_1] \quad (1.4)$$

The overall objective function of the cycle-consistent generative adversarial network(where $\lambda$ was the weight for cycle consistency loss):

$$\mathcal{L}(Gx\_y,Gy\_x,D_X,D_Y) = \mathcal{L}_{GAN}(Gx\_y,D_Y,X,Y)$$
$$+ \mathcal{L}_{GAN}(Gy\_x,D_X,Y,X)$$
$$+ \lambda\mathcal{L}_{cyc}(Gx\_y,Gy\_x)$$
$$+ \mathcal{L}_{idt}(Gx\_y,Gy\_x) \quad (1.5)$$

## 2.3 Network Structure

Eq.(1.5) is the Cyclegan overall objective loss function. We remove identity mapping to resist color sensitivity and speed up. But the generated graphics are seriously distorted. To avoid this, we change the number of layers of the residual network in the middle layer of the generator to increase the resolution, which makes the whole network training very slow.

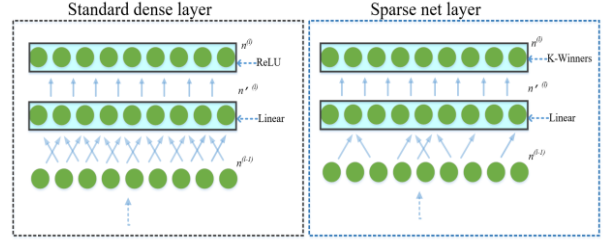

Fig 2. Comparison of dense and sparse layers. The weights of the sparse network are randomly sampled from a sparse subset of the underlying source layer. *n* represents the number of active units, and *l* represents the number of layers. *n l* indicates that the number of active units output by the *l* layer is *n*.

We decided to use a sparse network to replace the traditional dense network and replace the ReLU layer with sparse activation K-winners[19].In a sparse network, the number of non-zero products at each layer is approximately (sparseness of layer $l$) × (sparse weight of layer $l + 1$). This results in a differentiable sparse layer, which can be put into standard linear and convolutional layers[19]. The purpose is to reduce the weight parameters in the training process to achieve the effect of network pruning. Improves generator noise robustness and speeds up generator training. We use the random correction unit RReLU as an activation layer in the discriminator to improve the discriminator training stability.

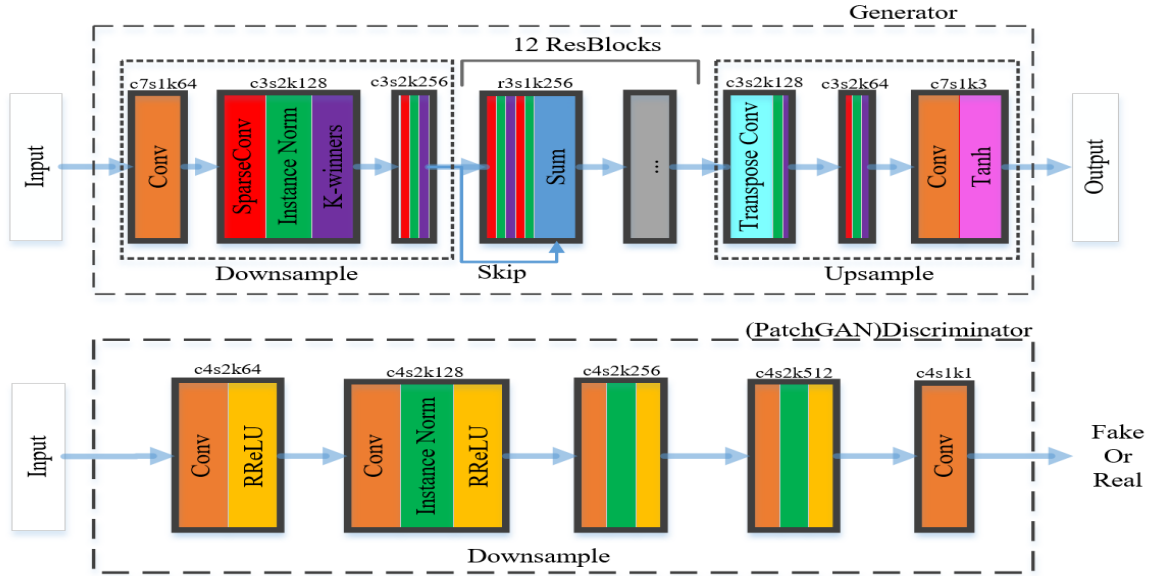

Fig 3. Our improved network (Fine-tuning). After removing the identity mapping function, the generator middle layer adds three residual blocks. All except decoders replace traditional dense network and ReLU activation layers with sparse convolution and sparse activation K-Winners. Let c7s1k64 denote a layer with 64 filters 7*7 convolution and stride 1. r is the residual. The discriminator still uses Patchgan, and we use the random correction unit RReLU as the discriminator activation layer.

## 3. EXPERIMENTAL

### 3.1 Training Details

We use GPU for training. Among them, 12 residual blocks are used in the middle layer of the network as the residual network [18]. The learning rate is set to 0.0002 during training. The result after optimizing D is divided by 2, which effectively slows down the learning rate of the discriminator. A total of 200 epochs were trained. The same learning rate is used for the first 100 epochs, and the learning rate decays linearly to zero for the last 100 epochs. The identity mapping function is removed, and the cycle consistency loss weight λ=10. The training weights are initialized from a Gaussian distribution N (0, 0.02). The Adam[25] optimizer with a second-order matrix decay rate of 0.999 was used. The original data set comes from the production atlas of a riveting and welding heavy-pressure equipment manufacturing company. As we all know, the data set affects the accuracy of the training model to a certain extent. We perform data enhancement on the original atlas. The engineering graphics data set contains 960 training sets and 240 test sets, and the size of the graphics after batch preprocessing is 256x256.

### 3.2 Cyclegan[16]

Using 1200 welding engineering drawing data sets for training and testing, it randomly selects 960 sample data sets to input into the network and trains epoch 200. It takes about 96 hours to get the trained network model.

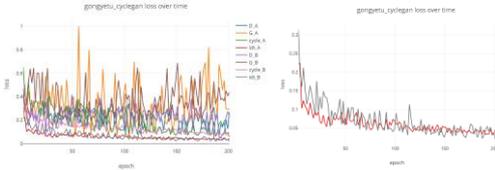

Fig 4. The loss curve shows that the cycle consistency loss and the identity map loss A and B converge stably around 0.05 (**left**). Its discriminators A and B learned well, and the model trained successfully (**right**). The overall model does not show model collapse and gradient explosion.

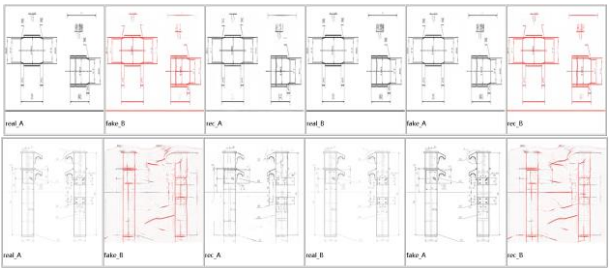

Fig 5. We are visualizing the training process. The integrity of the fake samples it generates is feasible. Part of the graphics is distorted, and the multi-line atlas is seriously distorted (the second line Fake B). When we change the color of the sample, it produces a fake red sample, which is not what we want.

Experiments show that when Cyclegan is first trained to translate welding structure engineering drawings, its translation effect on engineering drawings with different contents is different. It does not meet overall industrial manufacturing standards. We need to improve and optimize the network, but the initial training verified the feasibility of using Cyclegan to translate welding drawings

### 3.3 Cyclegan-Idt

The network model is trained with random 960 sample data. After epoch 200, the trained network model can be obtained in about 27 hours.

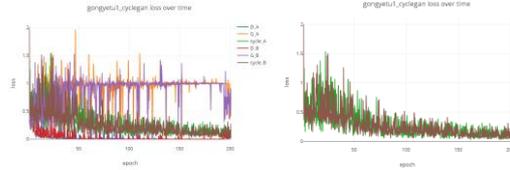

Fig 6. Generator A and B curves have been oscillating at 1.00 (**left**). Cycle Consistency A and B loss curves close to 0 (**right**). It can be seen that because the discriminator is too powerful, the learning ability of the generator is not good.

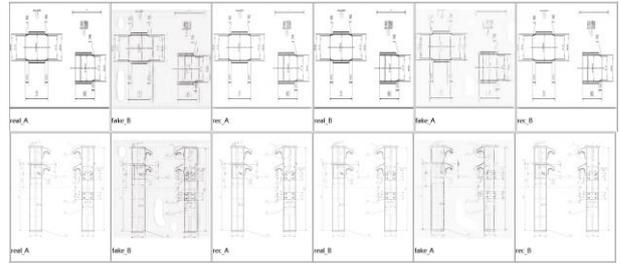

Fig 7. We were generating fake samples A and B with severe distortion. Large graphics distortion (Fake A and B) is present for both graphics with fewer lines and annotations or graphics with more lines and annotations. When we tried to add local color perturbations, the resulting samples did not change.

The translated content is recognized as welding engineering graphics, and there is not much requirement for the color difference between domains. Therefore, to improve the training speed of the model, we abandon the identity mapping function in CycleGAN to generate the adversarial network objective function:

$$\mathcal{L}(G_{x\_y}, G_{y\_x}, D_X, D_Y) = \mathcal{L}_{GAN}(G_{x\_y}, D_Y, X, Y) \\ + \mathcal{L}_{GAN}(G_{y\_x}, D_X, Y, X) \\ + \lambda \mathcal{L}_{cyc}(G_{x\_y}, G_{y\_x}) \quad (1.6)$$

After the network discards the identity mapping function, its training speed is significantly improved, which is about 3.5 times that of the traditional model. But from the results, it is far inferior to the traditional recurrent generative adversarial network as a whole. Although the training speed has been greatly improved, the advantages of fast speed far cannot make up for the shortcomings of accuracy.

### 3.4 Fine-Tuning(Ours)

Randomly select 960 sample data sets to train our network model. After epoch 200, it takes about 65 hours to get the trained network model.

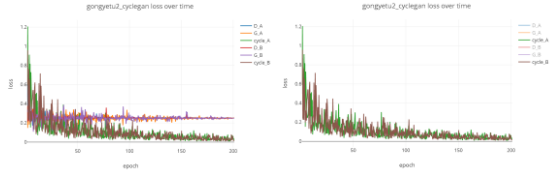

Fig 8. Generators A and B are relatively stable (**left**). Its discriminator can implement strong learning and eventually reach a Nash equilibrium (**right**). Its training converges at epoch 50, with faster convergence, stable structure, and no model collapse and exploding gradients.

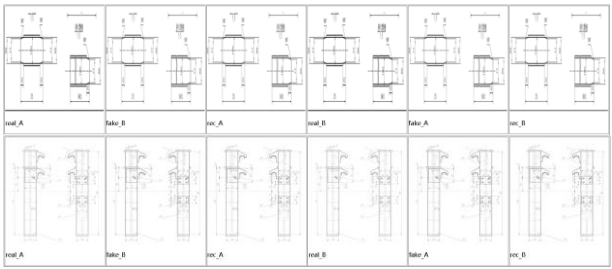

Fig 9. Whether translating sheet metal graphics with fewer lines and annotations or general engineering graphics with more line annotations, our network has greatly improved the quality of its translations. The generated fake samples A and B have high resolution and no severe distortion.

As shown in Fig 3. it is our improved network architecture. Our residual generator network has 12 residual blocks in the middle layer and uses a sparse network instead of a dense network in the encoder. Replacing the ReLU layer with K-Winners can increase the generator resolution and reduce the network weight parameters, and also increase the noise robustness of the generator. The discriminator activation layer uses RReLU as the activation layer to improve the stability of the discriminator training. It is a cycle-consistency-based generative adversarial network for translating welding engineering graphics.

Compare the training visualizations in Fig 5, Fig 7, and Fig 9. No serious graphics distortion. Our network can translate well engineering drawings with fewer lines and annotations, or drawings with more lines and annotations. It can meet the welding accuracy requirements in industrial manufacturing and has great advantages in speed, which is about 1.47 times that of traditional networks.

We qualitatively evaluate the C, C-idt, and C-idt+ fine-tuned network models using the Welded Structural Engineering Drawing dataset, respectively, as shown in Figure 10. We refer to the Cyclegan model as C for short in this paper. P is denoted as a color perturbation, i.e.the case where the training atlas is labeled colors. In Figure 10, the purple dashed box represents the annotations and mesh outlines that need to be translated into engineering drawings.

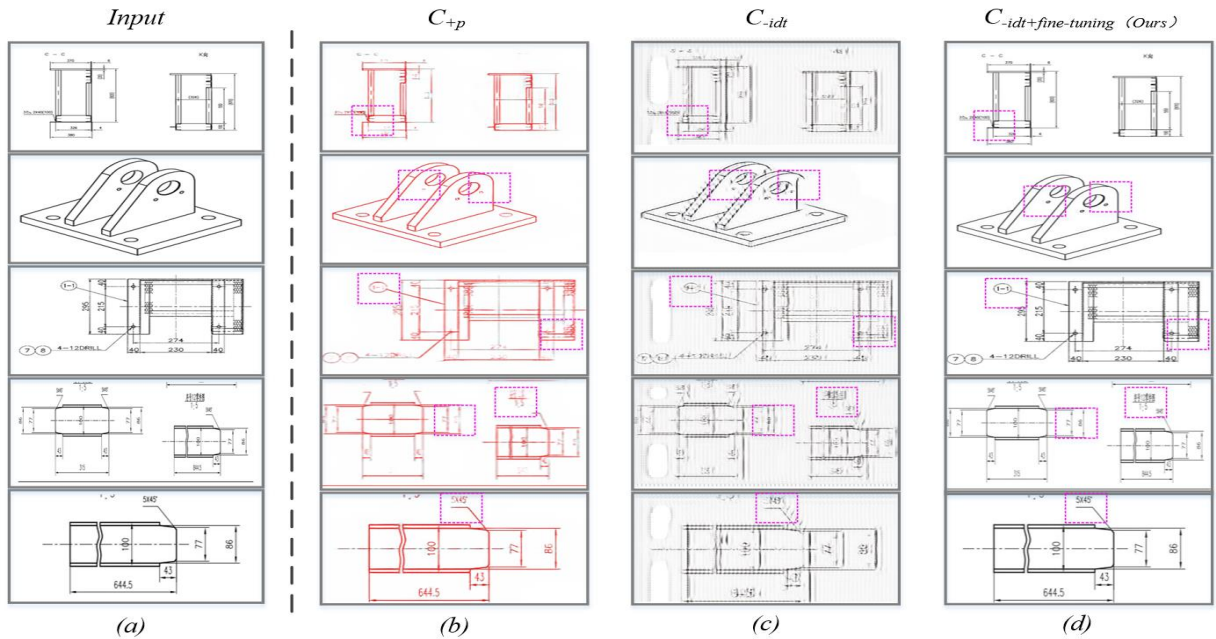

Fig 10. Method to compare experimental results. Column (a) is a drawing of the welded structure column (b) Cyclegan model translation effect after adding color perturbation (p). Column (c) is the translation effect of removing the identity mapping function. (d) The column is the translation effect of adding residual network and sparse algorithm (ours) based on the previous. The purple dotted box is the key translation object of the engineering drawing.

*3.5 Evaluation*

We evaluate the work with evaluation metrics SSIM, MSE, and PSNR[26][27][28] for image generation quality. SSIM (Structural Similarity) is used to evaluate the similarity level between two images. The range of SSIM is (0,1). For images X and Y, the closer the value of SSIM is to 1, the more similar the two images are. MSE (Mean Squared Error) mainly evaluates the degree of difference at the pixel level between the restored image i and the original image k. The smaller the MSE value, the more similar the two images are. PSNR (Peak Signal-to-Noise Ratio) is used to evaluate the distortion level and noise level of an image. The larger the PSNR, the less distortion of the generated image and the better the image quality.

Table 1: Cyclegan, demap, and our proposed network, training translation engineering drawing quality metrics comparison. Bold font means excellent.

| base | $\lambda_{idt}$ | fine-tune | Epoch | PSNR | SSIM(%) | MSE |
|---|---|---|---|---|---|---|
|  |  |  | 20 | 13.27 | 50.80 | 3063.37 |
| ✓ | ✓ | ✗ | 100 | 21.55 | 79.56 | **455.07** |
|  |  |  | 200 | **27.00** | **91.73** | 515.77 |
|  |  |  | 20 | 11.78 | 36.65 | 4311.68 |
| ✓ | ✗ | ✗ | 100 | **19.26** | **66.32** | **771.30** |
|  |  |  | 200 | 17.33 | 58.40 | 1202.53 |
|  |  |  | 20 | 29.96 | 97.11 | 65.666 |
| ✓ | ✗ | ✓ | 100 | 37.84 | 99.06 | 10.70 |
|  |  |  | 200 | **42.44** | **99.36** | **3.70** |

Table 2: Cyclegan, demap, and our proposed network test translation quality metrics comparison. Bold font means excellent.

| Base | $\lambda_{idt}$ | fine-tuning | PSNR | SSIM(%) | MSE |
|---|---|---|---|---|---|
| ✓ | ✓ | ✗ | 32.80 | 97.33 | 34.15 |
| ✓ | ✗ | ✗ | 18.00 | 60.90 | 1030.91 |
| ✓ | ✗ | ✓ | **44.89** | **99.58** | **2.11** |

We randomly test 240 welding industry drawings with a trained traditional, recurrent, adversarial network model. The graphs translated by Cyclegan (without adding color perturbation) have a graph structure similarity (SSIM) of up to 97.33% (retaining four significant figures). When it translates engineering drawing lines and welding drawings with many annotations, its mean square error (MSE) result can reach a minimum of 34.15. The graphics it translates are partially blurred and distorted; the overall graphics quality requirements for manufacturing cannot be met. But it confirmed the feasibility of Cyclegan to translate engineering graphics. Under the same conditions, we test the network model of discarding the identity mapping function. We can get the adversarial network that discards the identity mapping function from the results.

It does not matter whether it translates fewer lines and annotations or sheet metal engineering graphics with more lines and annotations; after translation, The graphics quality is greatly reduced. Its translated graph structure similarity (SSIM) is only 66.32% at maximum, and the mean square error (MSE) result is at least 771.30. Its translated engineering drawings are severely blurred and distorted. It cannot be used in industrial manufacturing. But its network structure is no longer sensitive to color.

Our network-translated Graph Structure Similarity (SSIM) improves by two percentage points over the Cyclegan model, reaching a maximum of 99.58%. Its mean square error (MSE) and peak signal-to-noise ratio (PSNR) have been improved from 34.15 and 32.80 to 2.11 and 44.89, respectively. It is about 1.47 times faster than Cyclegan under the same conditions. It can improve the noise robustness of the generator while ensuring the translation quality meets industrial manufacturing.

**4. CONCLUSION**

In this paper, we apply the Cyclegan network to welding engineering manufacturing. We found that using the sparse algorithm for the generator can greatly reduce the weight parameters of the generator and speed up network training. We replace the ReLU activation layer with a K-winners layer, which allows the active units n in each layer to have the same active frequency, which improves the noise robustness and training speed of the generator. In the discriminator, we still use the PatchGan discriminator, which uses a fully convolutional output N*N matrix to reflect the input image. Multiple patches correspond to the local features of the original image, which can better evaluate the images generated by the generator.

For time reasons, we are validating on a small dataset. Because adding intermediate layers greatly increases training time, we currently do not perform experiments on large datasets. Therefore, we are not sure whether this network also significantly affects the vast dataset. In the following work, the sparse algorithm can be applied to the new network containing VGGNet[29], which will be a good prospect.

**ACKNOWLEDGMENT**

This research was funded by the Advanced Manufacturing Institute of the School of Mechanical and Electrical Engineering, Xi'an Technological University, Shaanxi Provincial Key Science and Technology Project (Project No.:2022PT-02), and the Shaanxi Heavy Pressure Riveting and Welding Company.